\documentclass[11pt,letterpaper]{article}
\usepackage{naaclhlt2016}
\usepackage{times}
\usepackage{latexsym}

\naaclfinalcopy 
\def\naaclpaperid{***} 

\usepackage{url}
\usepackage{amsmath}
\usepackage{graphicx}
\usepackage{color}
\usepackage{subcaption}

\title{Recurrent Neural Network Encoder with Attention for Community Question Answering}

\author{Wei-Ning Hsu \and Yu Zhang \and James Glass\\
	    Computer Science and Artificial Intelligence Laboratory\\
	    Massachusetts Institute of Technology\\
	    Cambridge, MA 02139, USA\\
	    {\tt \{wnhsu,yzhang87,jrg\}@csail.mit.edu}
	    }

\date{}

\begin{document}
\maketitle

\begin{abstract}
We apply a general recurrent neural network (RNN) encoder framework to community question answering (cQA) tasks. Our approach does not rely on any linguistic processing, and can be applied to different languages or domains. Further improvements are observed when we extend the RNN encoders with a neural attention mechanism that encourages reasoning over entire sequences.  To deal with practical issues such as data sparsity and imbalanced labels, we apply various techniques such as transfer learning and multitask learning. Our experiments on the SemEval-2016 cQA task show 10\% improvement on a MAP score compared to an information retrieval-based approach, and achieve comparable performance to a strong handcrafted feature-based method.
\end{abstract}

\section{Introduction}
Community question answering (cQA) is a paradigm that provides forums for users to ask or answer questions on any topic with barely any restrictions. In the past decade, these websites have attracted a great number of users, and have accumulated a large collection of question-comment threads generated by these users. However, the low restriction results in a high variation in answer quality, which makes it time-consuming to search for useful information from the existing content. It would therefore be valuable to automate the procedure of ranking related questions and comments for users with a new question, or when looking for solutions from comments of an existing question.

Automation of cQA forums can be divided into three tasks: question-comment relevance (Task A), question-question relevance (Task B), and question-external comment relevance (Task C). One might think that classic retrieval models like language models for information retrieval \cite{LMIR} could solve these tasks. However, a big challenge for cQA tasks is that users are used to expressing similar meanings with different words, which creates gaps when matching questions based on common words. Other challenges include informal usage of language, highly diverse content of comments, and variation in the length of both questions and comments.

To overcome these issues, most previous work (e.g. SemEval 2015~\cite{SemEval2015}) relied heavily on additional features and reasoning capabilities. In~\cite{rocktaschel2015reasoning}, a neural attention-based model was proposed for automatically recognizing entailment relations between pairs of natural language sentences. In this study, we first modify this model for all three cQA tasks. We also extend this framework into a jointly trained model when the external resources are available, i.e. selecting an external comment when we know the question that the external comment answers (Task C). 

Our ultimate objective is to classify relevant questions and comments without complicated handcrafted features. By applying RNN-based encoders, we avoid heavily engineered features and learn the representation automatically. In addition, an attention mechanism augments encoders with the ability to attend to past outputs directly. This becomes helpful when encoding longer sequences, since we no longer need to compress all information into a fixed-length vector representation. 

In our view, existing annotated cQA corpora are generally too small to properly train an end-to-end neural network. To address this, we investigate transfer learning by pretraining the recurrent systems on other corpora, and also generating additional instances from existing cQA corpus.

\section{Related Work}
Earlier work of community question answering relied heavily on feature engineering, linguistic tools, and external resource. \cite{jeon2006framework} and \cite{shah2010evaluating} utilized rich non-textual features such as answer's profile. \cite{grundstrom2014using} syntactically analyzed the question and extracted name entity features. \cite{harabagiu2006methods} demonstrated a textual entailment system can enhance cQA task by casting question answering to logical entailment.

More recent work incorporated word vector into their feature extraction system and based on it designed different distance metric for question and answer \cite{tran2015jaist} \cite{VectorSLU}. While these approaches showed effectiveness, it is difficult to generalize them to common cQA tasks since linguistic tools and external resource may be restrictive in other languages and features are highly customized for each cQA task.

Very recent work on answer selection also involved the use of neural networks. \cite{WangRNNCQA} used LSTM to construct a joint vector based on both the question and the answer and then converted it into a learning to rank problem. \cite{feng2015applying} proposed several convolutional neural network (CNN) architectures for cQA. Our method differs in that RNN encoder is applied here and by adding attention mechanism we jointly learn which words in question to focus and hence available to conduct qualitative analysis. During classification, we feed the extracted vector into a feed-forward neural network directly instead of using mean/max pooling on top of each time steps.

\section{Method}
\begin{figure*}[!ht]
\centering
\includegraphics[width=0.8\textwidth]{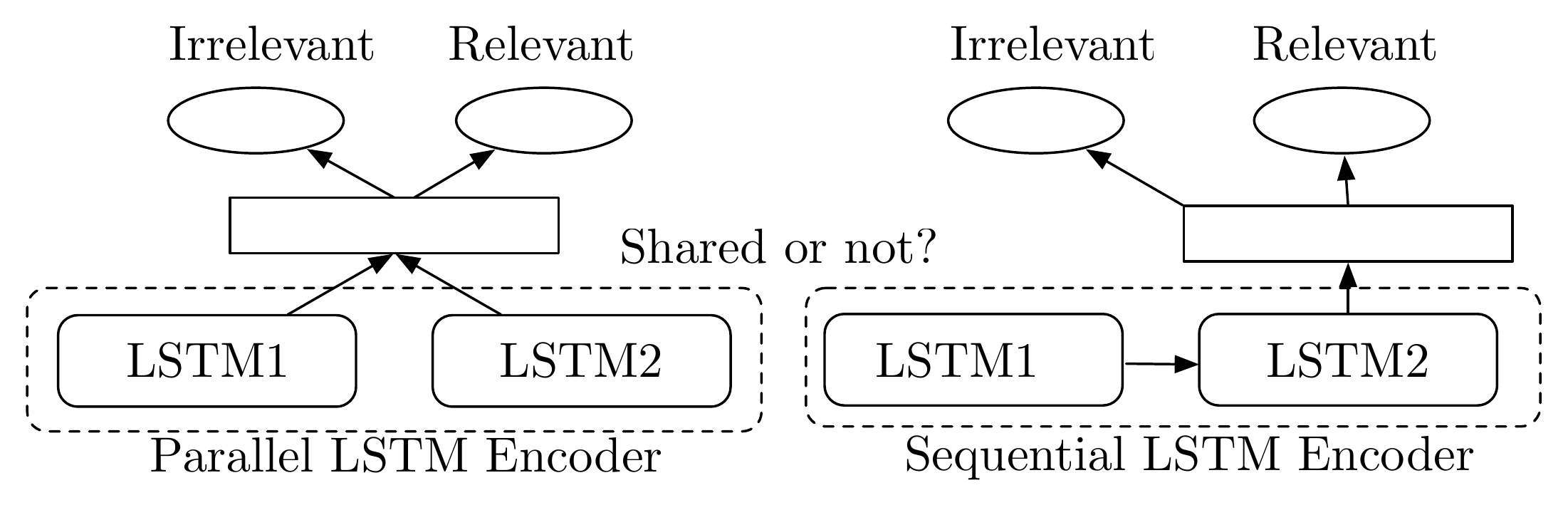}
\caption{RNN encoder for related question/comment selection.}
\label{fig:parmodel}
\end{figure*}
In this section, we first discuss long short-term memory (LSTM) units and an associated attention mechanism.  Next, we explain how we can encode a pair of sentences into a dense vector for predicting relationships using an LSTM with an attention mechanism. Finally, we apply these models to predict question-question similarity, question-comment similarity, and question-external comment similarity.

\subsection{LSTM Models}
LSTMs have shown great success in many different fields. An LSTM unit contains a memory cell with self-connections, as well as three multiplicative gates to control information flow.  Given input vector $x_t$, previous hidden outputs $h_{t-1}$, and previous cell state $c_{t-1}$, LSTM units operate as follows:
\begin{align}
X &= \begin{bmatrix}
    x_t\\[0.3em]
    h_{t-1}\\[0.3em]
    \end{bmatrix}\\
i_t &= \sigma(\mathbf{W_{iX}}X + \mathbf{W_{ic}}c_{t-1} + \mathbf{b_i})\\
f_t &= \sigma(\mathbf{W_{fX}}X + \mathbf{W_{fc}}c_{t-1} + \mathbf{b_f})\\
o_t &= \sigma(\mathbf{W_{oX}}X + \mathbf{W_{oc}}c_{t-1} + \mathbf{b_o})\\
c_t &= f_t \odot c_{t-1} + i_t \odot tanh(\mathbf{W_{cX}}X + \mathbf{b_c})\\
h_t &= o_t \odot tanh(c_t)
\end{align}

\noindent where $i_t$, $f_t$, $o_t$ are input, forget, and output gates, respectively. The sigmoid function $\sigma()$ is a soft gate function controlling the amount of information flow. $W$s and $b$s are model parameters to learn.

\subsection{Neural Attention}

A traditional RNN encoder-decoder approach~\cite{sutskever2014sequence} first encodes an arbitrary length input sequence into a fixed-length dense vector that can be used as input to subsequent classification models, or to initialize the hidden state of a secondary decoder. However, the requirement to compress all necessary information into a single fixed length vector can be problematic.  A neural attention model~\cite{bahdanau2014neural} \cite{cho2014learning} has been recently proposed to alleviate this issue by enabling the network to attend to past outputs when decoding. Thus, the encoder no longer needs to represent an entire sequence with one vector; instead, it encodes information into a sequence of vectors, and adaptively chooses a subset of the vectors when decoding.

\begin{figure*}[!ht]
\centering
\includegraphics[width=0.8\textwidth]{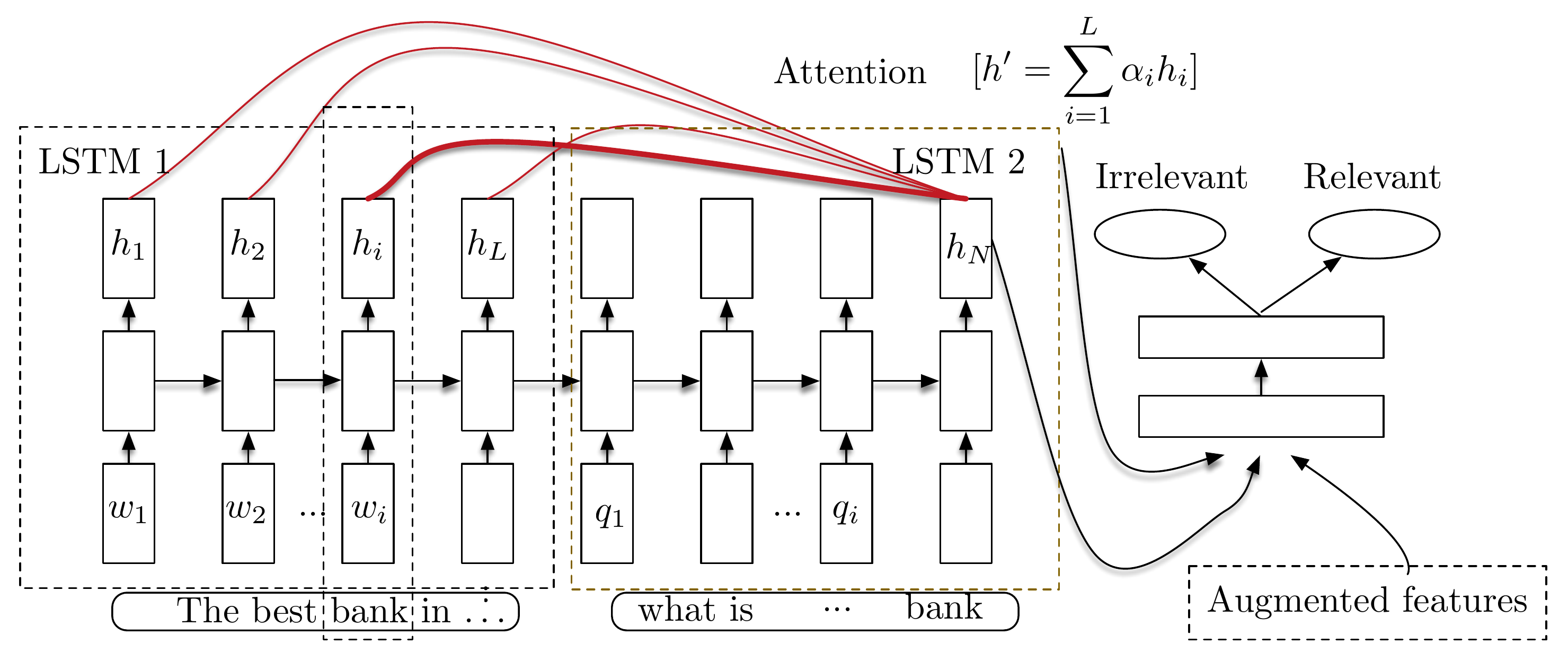}
\caption{Neural attention model for related question/comment selection.}
\label{fig:attention}
\end{figure*}

\subsection{Predicting Relationships of Object Pairs with an Attention Model}
In our cQA tasks, the pair of objects are (question, question) or (question, comment), and the relationship is relevant/irrelevant.  The left side of Figure~\ref{fig:parmodel} shows one intuitive way to predict relationships using RNNs. Parallel LSTMs encode two objects independently, and then concatenate their outputs as an input to a feed-forward neural network (FNN) with a softmax output layer for classification.

The representations of the two objects are generated independently in this manner. However, we are more interested in the relationship instead of the object representations themselves. Therefore, we consider a serialized LSTM-encoder model in the right side of Figure~\ref{fig:parmodel} that is similar to that in~\cite{rocktaschel2015reasoning}, but also allows an augmented feature input to the FNN classifier.

Figure~\ref{fig:attention} illustrates our attention framework in more detail. The first LSTM reads one object, and passes information through hidden units to the second LSTM. The second LSTM then reads the other object and generates the representation of this pair after the entire sequence is processed.  We build another FNN that takes this representation as input to classify the relationship of this pair.

By adding an attention mechanism to the encoder, we allow the second LSTM to attend to the sequence of output vectors from the first LSTM, and hence generate a weighted representation of first object according to both objects. Let $h_N$ be the last output of second LSTM and $M = [h_1, h_2, \cdots, h_L]$ be the sequence of output vectors of the first object. The weighted representation of the first object is
\begin{equation}
h' = \sum_{i=1}^{L} \alpha_i h_i
\end{equation}
The weight is computed by
\begin{equation}
\alpha_i = \dfrac{exp(a(h_i,h_N))}{\sum_{j=1}^{L}exp(a(h_j,h_N))}
\end{equation}
where $a()$ is the importance model that produces a higher score for $(h_i, h_N)$ if $h_i$ is useful to determine the object pair's relationship. We parametrize this model using another FNN. Note that in our framework, we also allow other augmented features (e.g., the ranking score from the IR system) to enhance the classifier. So the final input to the classifier will be $h_N$, $h'$, as well as augmented features.

\subsection{Modeling Question-External Comments}
\label{sec:multi}
For task C, in addition to an original question (oriQ) and an external comment (relC), the question which relC commented on is also given (relQ). To incorporate this extra information, we consider a multitask learning framework which jointly learns to predict the relationships of the three pairs (oriQ/relQ, oriQ/relC, relQ/relC).

\begin{figure*}[!t]
\centering
\includegraphics[width=0.8\textwidth]{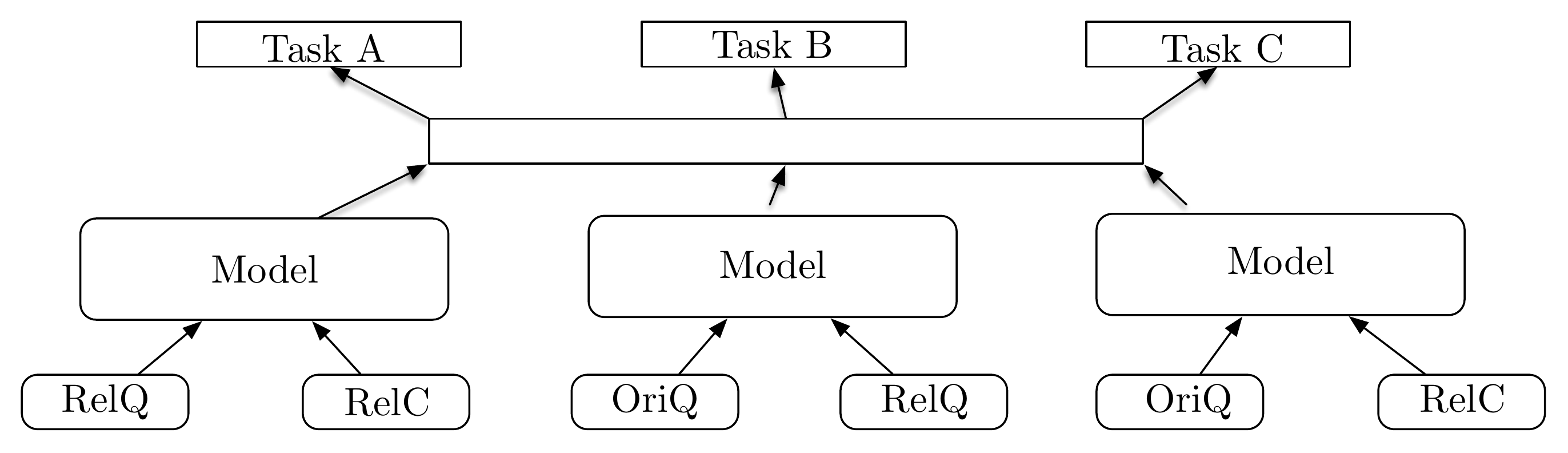}
\caption{Joint learning for external comment selection.}
\label{fig:multi}
\end{figure*}
Figure~\ref{fig:multi} shows our framework: the three lower models are separate serialized LSTM-encoders for the three respective object pairs, whereas the upper model is an FNN that takes as input the concatenation of the outputs of three encoders, and predicts the relationships for all three pairs. More specifically, the output layer consists of three softmax layers where each one is intended to predict the relationship of one particular pair.

For the overall loss function, we combine three separate loss functions using a heuristic weight vector $\beta$ that allocates a higher weight to the main task (oriQ-relC relationship prediction) as follows:
\begin{equation}
\mathcal{L} = \beta_1 \mathcal{L}_1 + \beta_2 \mathcal{L}_2 + \beta_3 \mathcal{L}_3
\end{equation}
By doing so, we hypothesize that the related tasks can improve the main task by leveraging commonality among all tasks.

\section{Experiments}

We evaluate our approach on all three cQA tasks. We use the cQA datasets provided by the Semeval 2016 task~\footnote{http://alt.qcri.org/semeval2016/task3}. The cQA data is organized as follows: there are 267 original questions, each question has 10 related question, and each related question has 10 comments. Therefore, for task A, there are a total number of 26,700 question-comment pairs. For task B, there are 2,670 question-question pairs. For task C, there are 26,700 question-comment pairs. The test dataset includes 50 questions, 500 related questions and 5,000 comments which do not overlap with the training set. To evaluate the performance, we use mean average precision (MAP) and F1 score.

\begin{figure*}[!ht]
\centering
\includegraphics[width=0.8\textwidth]{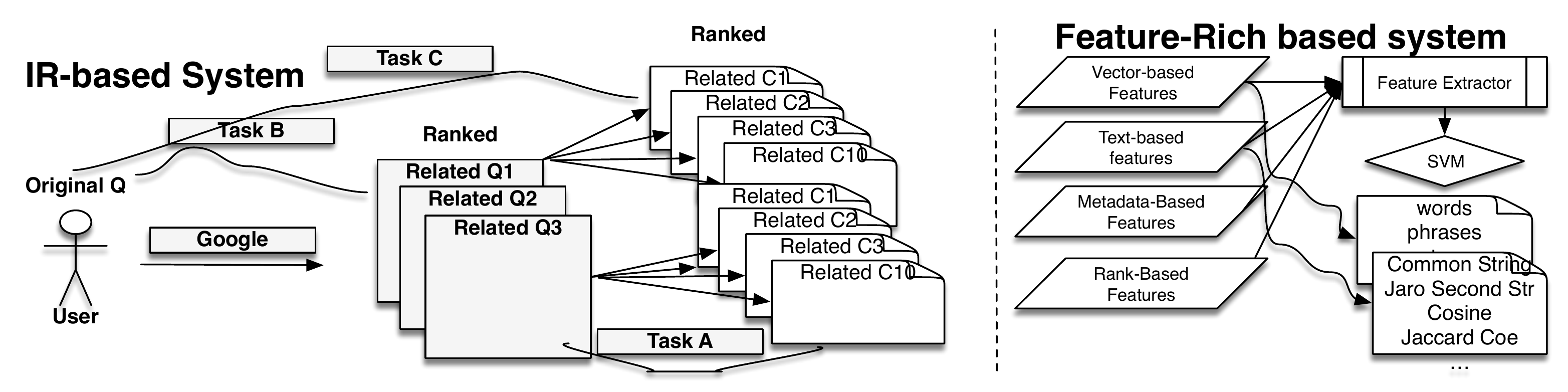}
\caption{IR-based system and feature-rich based system.}
\label{fig:baseline}
\end{figure*}

\paragraph{Baseline System:} Figure~\ref{fig:baseline} illustrates our baseline systems. The IR-based system is scored by the Google search engine. For each question-comment pair, or question-question pair, we use Google's rank to calculate the MAP.  While there is no training on the target data, we expect that Google used many external resources to produce these ranks. The feature-rich system is that proposed by \cite{VectorSLU} in SemEval-2015. In this approach, they compute \textit{text-based}, \textit{vector-based}, \textit{metadata-based} and \textit{rank-based features} from the pre-processed data. The features are used by a linear SVM for comment selection. This system includes traditional handcrafted features, and some RNN-based features (word vectors). It also includes the information from the IR system (ranked-based). So we believe it is a strong baseline to compare with our model.

\paragraph{RNN encoder:} Our system is based on Theano~\cite{Bastien-Theano-2012,bergstra+al:2010-scipy}. Table~\ref{tab:hyper} gives a list of hyper-parameters we considered.  As suggested by~\cite{DBLP:journals/corr/GreffSKSS15}, the hyper-parameters for LSTMs can be tuned independently. We tuned each parameter separately on a development set (split from the training set) and simply picked the best setting. Our experiments show that using word embeddings from Google-News provides modest improvements, but fixing the embedding degrades performance a lot. Also, using separate parameters for LSTMs is better than sharing. For the optimization method, AdaDelta converged faster, but AdaGrad gives better performance. Note that all the parameters were tuned on Task A, and we simply applied them to Task B and C. This is for saving computation, and also because Task A is more well-defined compared to B and C in terms of dataset size and label balance.
\begin{table}[!ht]
\centering
\begin{tabular}{c|c}
Embedding & \textbf{init} or random, fix or \textbf{update} \\
 Two LSTM  & shared or \textbf{not} \\
 \#cells for LSTM & 64, \textbf{128}, 256\\
 \# nodes for MLP & 128, \textbf{256}\\
 Optimizer & \textbf{AdaGrad}, AdaDelta, SGD\\
 learning rate & 0.001,\textbf{0.01},0.1\\
 Regularizer & \textbf{Dropout}, L2 regularization\\
 Dropout rate & 0.0, 0.2, 0.3, \textbf{0.4}, 0.5\\
 L2 & 0, 0.001, \textbf{0.0001}, 0.00001\\
\end{tabular}
\caption{The hyper-parameters we tuned. Terms in bold represent the selected final parameters.}
\label{tab:hyper}
\end{table}

\subsection{Preliminary Results}
\begin{table*}[!ht]
\centering
\begin{tabular}{|c|c|c|c|c|c|c|}
\hline
&\multicolumn{2}{|c|}{Task A} & \multicolumn{2}{|c|}{Task B} & \multicolumn{2}{|c|}{Task C}\\\hline\hline 
 Model & MAP & F1 & MAP & F1 & MAP & F1\\\hline
 Random & 0.4860 & 0.5004 & 0.5595 & \bf{0.4691} & 0.1383 & 0.1277\\
 Parallel LSTM & 0.6123 & 0.6091 & 0.5553 & 0.4087 & 0.2413 & 0.0057\\
 Seq LSTM & 0.6175 & 0.6063 & 0.5620 & 0.4299 & 0.2356 & 0.0115\\
 w/ Attention & \bf{0.6239} & \bf{0.6323} & \bf{0.5723} & 0.4334 & \bf{0.2837} & \bf{0.1449}\\\hline
\end{tabular}
\caption{The RNN encoder results for cQA tasks (bold is best).}
\label{tab:preResults}
\end{table*}

Table~\ref{tab:preResults} shows the initial results using the RNN encoder for different tasks. We observe that the attention model always gets better results than the RNN without attention, especially for task C. However, the RNN model achieves a very low F1 score. For task B, it is even worse than the random baseline. We believe the reason is because for task B, there are only 2,670 pairs for training which is very limited training for a reasonable neural network. For task C, we believe the problem is highly imbalanced data. Since the related comments did not directly comment on the original question, more than $90\%$ of the comments are labeled as irrelevant to the original question. The low F1 (with high precision and low recall) means our system tends to label most comments as irrelevant. In the following section, we investigate methods to address these issues.

\subsection{Robust Parameter Initialization}
One way to improve models trained on limited data is to use external data to pretrain the neural network. We therefore considered two different datasets for this task. 
\begin{itemize}
\item Cross-domain: The Stanford natural language inference (SNLI) corpus~\cite{SNLI} has a huge amount of cleaned premise and hypothesis pairs. Unfortunately the pairs are for a different task. The relationship between the premise and hypothesis may be similar to the relation between questions and comments, but may also be different.
\item In-domain: since task A seems has reasonable performance, and the network is also well-trained, we could use it directly to initialize task B.
\end{itemize}
To utilize the data, we first trained the model on each auxiliary data (SNLI or Task A) and then removed the softmax layer. After that, we retrain the network using the target data with a softmax layer that was randomly initialized.

For task A, the SNLI cannot improve MAP or F1 scores. Actually it slightly hurts the performance. We surmise that it is probably because the domain is different. Further investigation is needed: for example, we could only use the parameter for embedding layers etc.  For task B, the SNLI yields a slight improvement on MAP ($0.2\%$), and Task A could give ($1.2\%$) on top of that. No improvement was observed on F1.  For task C, pretraining by task A is also better than using SNLI (task A is $1\%$ better than the baseline, while SNLI is almost the same). 

In summary, the in-domain pretraining seems better, but overall, the improvement is less than we expected, especially for task B, which only has very limited target data. We will not make a conclusion here since more investigation is needed.

\subsection{Multitask Learning}
As mentioned in Section~\ref{sec:multi}, we also explored a multitask learning framework that jointly learns to predict the relationships of all three tasks. We set $0.8$ for the main task (task C) and $0.1$ for the other auxiliary tasks. The MAP score did not improve, but F1 increases to $0.1617$.  We believe this is because other tasks have more balanced labels, which improves the shared parameters for task C.

\subsection{Augmented data}
There are many sources of external question-answer pairs that could be used in our tasks. For example: WebQuestion (was introduced by the authors of SEMPRE system~\cite{webq}) and The SimpleQuestions dataset \footnote{\url{http://fb.ai/babi}.}.
All of them are positive examples for our task and we can easily create negative examples from it. Initial experiments indicate that it is very easy to overfit these obvious negative examples.  We believe this is because our negative examples are non-informative for our task and just introduce noise.  

Since the external data seems to hurt the performance, we try to use the in-domain pairs to enhance task B and task C.  For task B, if relative question $1$ (rel1) and relative question $2$ (rel2) are both relevant to the original question, then we add a positive sample (rel1, rel2, 1). If either rel1 and rel2 is irrelevant and the other is relevant, we add a negative sample (rel1, rel2, 0). After doing this, the samples of task B increase from $2,670$ to $11,810$. By applying this method, the MAP score increased slightly from $0.5723$ to $0.5789$ but the F1 score improved from $0.4334$ to $0.5860$.

For task C, we used task A's data directly. The results are very similar with a slight improvement on MAP, but large improvement on F1 score from $0.1449$ to $0.2064$.

\subsection{Augmented features}
 To further enhance the system, we incorporate a one hot vector of the original IR ranking as an additional feature into the FNN classifier. Table~\ref{tab:aug} shows the results. In comparing the models with and without augmented features, we can see large improvement for task B and C. The F1 score for task A degrades slightly but MAP improves. This might be because task A already had a substantial amount of training data.
\begin{table*}[!ht]
\centering
\begin{tabular}{|c|c|c|c|c|c|c|}
\hline
&\multicolumn{2}{|c|}{Task A} & \multicolumn{2}{|c|}{Task B} & \multicolumn{2}{|c|}{Task C}\\\hline\hline 
 Model & MAP & F1 & MAP & F1 & MAP & F1\\\hline
 w/ Attention & 0.6239 & {\bf 0.6323} & 0.5723 & 0.4334 & 0.2837 & 0.1449\\\hline
 w/ Attention + aug features & {\bf 0.6385} & 0.6218 & {\bf 0.6585} & {\bf 0.5382} & {\bf 0.3236} & {\bf 0.1963}\\\hline
\end{tabular}
\caption{cQA task results with augmented features (bold is best).}
\label{tab:aug}
\end{table*}

\subsection{Comparison with Other Systems}
Table~\ref{tab:final} gives the final comparison between different models (we only list the MAP score because it is the official score for the challenge). Since the two baseline models did not use any additional data, in this table our system was also restricted to the provided training data. For task A, we can see that if there is enough training data our single system already performs better than a very strong feature-rich based system. For task B, since only limited training data is given, both feature-rich based system and our system are worse than the IR system. For task C, our system also got comparable results with the feature-rich based system. If we do a simple system combination (average the rank score) between our system and the IR system, the combined system will give large gains on tasks B and C\footnote{The feature-rich based system was already combined with the IR system)}. This implies that our system is complimentary with the IR system. 

\begin{table}[!ht]
\centering
\begin{tabular}{|c|c|c|c|c|c|c|}
\hline
&\multicolumn{1}{|c|}{Task A} & \multicolumn{1}{|c|}{Task B} & \multicolumn{1}{|c|}{Task C}\\\hline\hline 
 Model & MAP  & MAP & MAP \\\hline
 IR & 0.538 & 0.714 & 0.307\\\hline
 Attention & 0.639 & 0.659 & 0.324\\\hline
 Feature-Rich \& IR & 0.632 & 0.685 & 0.339\\\hline
 Attention \& IR & {\bf 0.639} & {\bf 0.717} & {\bf 0.394}\\\hline
\end{tabular}
\caption{Compared with other systems (bold is best).}
\label{tab:final}
\end{table}

\section{Analysis of Attention Mechanism}
In addition to quantitative analysis, it is natural to qualitatively evaluate the performance of the attention mechanism by visualizing the weight distribution of each instance. We randomly picked several instances from the test set in task A, for which the sentence lengths are more moderate for demonstration. These examples are shown in Figure~\ref{fig:vis}, and categorized into short, long, and noisy sentences for discussion. A darker blue patch refers to a larger weight relative to other words in the same sentence.

\subsection{Short Sentences}
Figure~\ref{fig:short} illustrates two cQA examples whose questions are relatively short. The comments corresponding to these questions are \textit{``...snorkeling two days ago off the coast of dukhan...''} and \textit{``the doha international airport...''}. We can observe that our model successfully learns to focus on the most representative part of the question pertaining to classifying the relationship, which is \textit{"place for snorkeling"} for the first example and \textit{``place can ... visited in qatar''} for the second example.

\subsection{Long Sentences}
In Figure~\ref{fig:long}, we investigate two examples with longer questions, which both contain 63 words. Interestingly, the distribution of weights does not become more uniform; the model still focuses attention on a small number of hot words, for example, \textit{``puppy dog for ... mall''} and \textit{``hectic driving in doha ... car insurance ... quite costly''}. Additionally, some words that appear frequently but carry little information for classification are assigned very small weights, such as \textit{I/we/my}, \textit{is/am}, \textit{like}, and \textit{to}.

\subsection{Noisy Sentence}
\begin{figure*}[!t]
\begin{subfigure}{\textwidth}
    \centering
    \includegraphics[width=\textwidth]{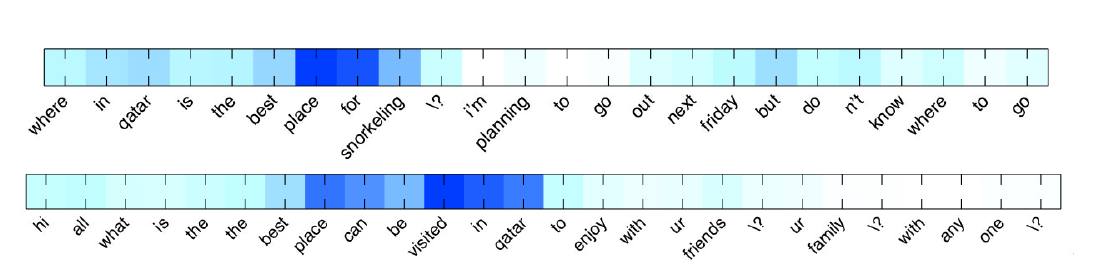}
    \caption{short sentences}
    \label{fig:short}
\end{subfigure}
\begin{subfigure}{\textwidth}
    \centering
    \includegraphics[width=\textwidth]{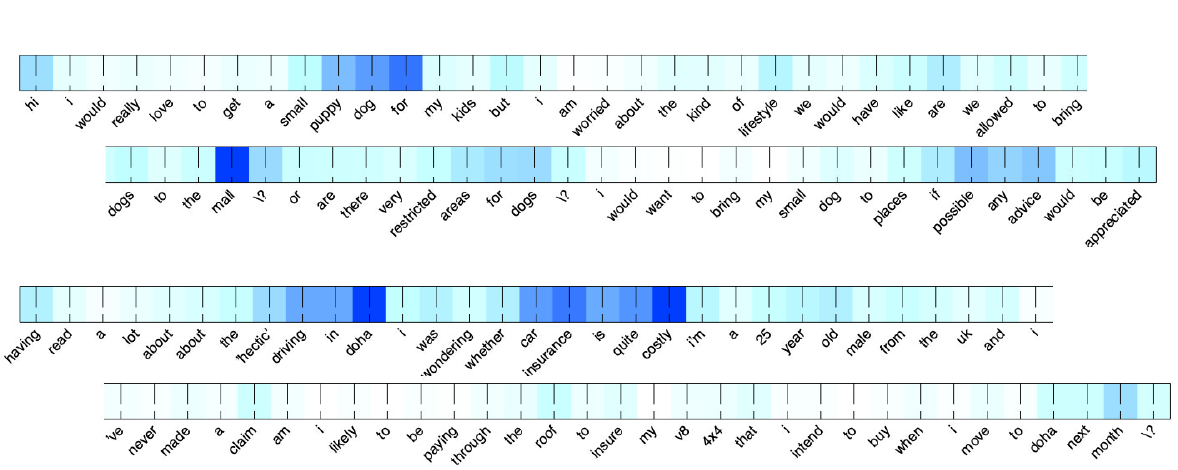}
    \caption{long sentences}
    \label{fig:long}
\end{subfigure}
\begin{subfigure}{\textwidth}
    \centering
    \includegraphics[width=\textwidth]{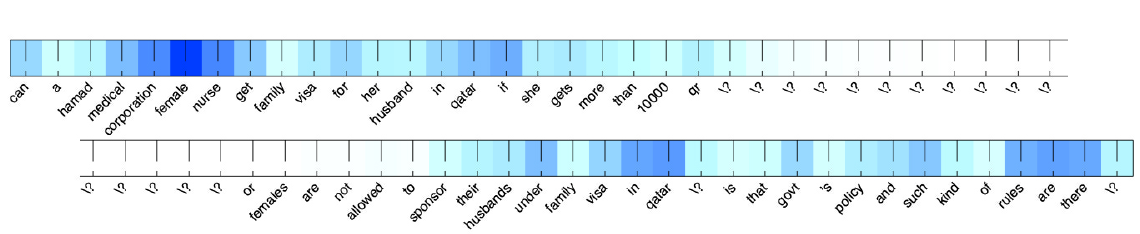}
    \caption{noisy sentence}
    \label{fig:noisy}
\end{subfigure}
\caption{Visualization of attention mechanism on short, long, and noisy sentences.}
\label{fig:vis}
\end{figure*}

Due to the open nature of cQA forums, some content is noisy. Figure~\ref{fig:noisy} is an example with excessive usage of question marks. Again, our model exhibits its robustness by allocating very low weights to the noise symbols and therefore excludes the noninformative content.

\section{Conclusion}
In this paper, we demonstrate that a general RNN encoder framework can be applied to community question answering tasks. By adding a neural attention mechanism, we showed quantitatively and qualitatively that attention can improve the RNN encoder framework.  To deal with a more realistic scenario, we expanded the framework to incorporate metadata as augmented inputs to a FNN classifier, and pretrained models on larger datasets,  increasing both stability and performance. Our model is consistently better than or comparable to a strong feature-rich baseline system, and is superior to an IR-based system when there is a reasonable amount of training data.

Our model is complimentary with an IR-based system that uses vast amounts of external resources but trained for general purposes. By combining the two systems, it exceeds the feature-rich and IR-based 
system in all three tasks.

Moreover, our approach is also language independent. We have also performed preliminary experiments on the Arabic portion of the SemEval-2016 cQA task. The results are competitive with a hand-tuned strong baseline from SemEval-2015. 

Future work could proceed in two directions: first, we can enrich the existing system by incorporating available metadata and preprocessing data with morphological normalization and out-of-vocabulary mappings; second, we can reinforce our model by carrying out word-by-word and history-aware attention mechanisms instead of attending only when reading the last word.

\bibliography{submission}
\bibliographystyle{naaclhlt2016}

\end{document}